# Zero Hallucination, by Construction: Hallucination-Aware Layered Oversight for Trustworthy Enterprise AI

Bogdan Răduță · Horia Velicu · Alexandru Preda · Șerban Chiricescu
*FlowX.AI*
bogdan.raduta@flowx.ai

**Abstract—** Enterprises will not deploy AI agents they cannot trust, and the most-cited reason for distrust is hallucination: confident, fluent output that is simply not true. The common response is to wait for a model that does not hallucinate. We argue that this is the wrong target. Large language models are, by construction, capable of generating unsupported text, and no amount of scale removes the possibility; a faithfulness judge bolted onto a raw model catches some errors but still ships others, and even well-curated retrieval pipelines have been shown to fabricate citations[1]. We reframe the goal: "zero hallucination" is not a property a model possesses but a property a system enforces. We present **HALO (Hallucination-Aware Layered Oversight)**, an assurance architecture which treats hallucination as a containable failure mode rather than an eliminable one. HALO composes six layers of defense: grounded generation over retrieved, approved content; constrained, deterministic execution that bounds where the model can err; multi-signal verification that scores every output for groundedness and hallucination using both an LLM judge and evidence-based checks against the source text; calibrated abstention, so the system declines rather than guesses when grounding is insufficient; total traceability of every retrieval, tool call, and generation; and continuous oversight that detects drift, alerts on threshold breaches, and closes the loop by regenerating and statistically validating improved agents. We detail each layer, give particular attention to evidence-based confidence (which verifies extractions against the source document rather than trusting the model's self-reported certainty), and illustrate the architecture on a regulated claims-extraction workload.

## 1 Introduction

Ask an enterprise risk officer why a promising AI agent never reached production, and the answer is rarely about latency or cost. It is about trust, and trust founders on a single failure mode: hallucination, the tendency of a language model to produce fluent, confident text that is not supported by any source[2]. In a consumer chatbot a hallucination is an annoyance. In a bank's loan-decisioning workflow or an insurer's claims pipeline it is a compliance incident, a mis-paid claim, or a fabricated citation in a regulated filing.

The instinct of the field has been to treat hallucination as a defect to be engineered out of the model: better pre-training, larger context, retrieval grounding, a faithfulness judge that scores the answer. Each of these helps, and none of them finishes the job. Retrieval-augmented generation reduces unsupported claims but does not eliminate them; a Stanford evaluation of leading legal-research tools found that systems marketed as hallucination-free still fabricated citations and misstated holdings[1]. Atomic-claim faithfulness scoring[3],[4] and self-consistency checks[5] catch many errors, but they are probabilistic detectors operating on a stochastic generator, and a detector with any false-negative rate will, at enterprise volume, ship unsupported answers. Worse, the most convenient confidence signal, the model's own self-reported certainty, is itself a model output and is poorly calibrated[6]: a hallucinated answer often arrives with high stated confidence.

We take a different stance. A large language model is, by its nature, a system that can generate unsupported text on any input; this is not a bug that scale will retire but a property of how the technology works. The right engineering question is therefore not "how do we build a model that cannot hallucinate?" but "how do we build a *system* in which a hallucination cannot reach the user unnoticed, unflagged, or uncorrected?” This is the same move safety engineering makes everywhere else: we do not demand a component that never fails, we design an architecture that contains the failure.

> *Zero hallucination is not a property a model possesses. It is a property a system enforces. The model is allowed to be fallible; the harness is not allowed to let that fallibility through unseen.*

This paper presents **HALO (Hallucination-Aware Layered Oversight)**, an assurance architecture and its monitoring plane. HALO does not claim a model that never errs. It claims something more defensible and more useful to a regulated enterprise: a layered system in which generation is grounded in approved content, execution is constrained so the model cannot wander, every output is verified against its sources by independent signals, the system abstains rather than guesses when evidence is thin, every step is traced, and quality is continuously monitored so that regressions are caught and corrected. “Zero hallucination” names the commitment that results: not the absence of a fallible component, but the presence of a harness that makes the failure observable, bounded, and actionable.

### 1.1 Contributions

- A reframing of the hallucination problem from model defect to system property, and a containment architecture, HALO, that operationalizes it as defense-in-depth rather than a single detector.
- A description of six composed layers (grounded generation, constrained execution, multi-signal verification, calibrated abstention, total traceability, continuous oversight) and how each closes a gap the others leave open.
- An emphasis on *evidence-based confidence*: verifying extracted values against the source document deterministically, rather than trusting the model’s self-reported certainty, which is itself a hallucinable output.
- A description of the oversight loop that detects drift on multi-metric baselines and closes the loop by regenerating and statistically validating improved agents before promotion.
- An illustrative regulated claims-extraction case study showing how the architecture contains and then corrects a hallucination regression.

## 2 Why a Single Detector Is Not Enough

### 2.1 The state of the practice

The mainstream 2026 pattern for trustworthy generation is retrieval-augmented generation paired with a faithfulness judge that gates the answer: retrieve context, generate, then score the output for groundedness and, if it scores below a threshold, re-retrieve or refuse. Frameworks such as RAGAS decompose an answer into atomic claims and verify each against the retrieved context[3]; FACTSCORE applies the same atomic-precision idea to long-form text[4]; SelfCheckGPT exploits sampling inconsistency as a black-box hallucination signal[5]. These are real advances, and HALO uses techniques from all of them.

### 2.2 Three gaps a single judge leaves open

A lone faithfulness check, however good, leaves predictable gaps. **First, detector error compounds with volume.** A judge with a 95% catch rate still passes one unsupported answer in twenty; at the throughput of a production claims pipeline that is thousands of unflagged errors a month. A single layer cannot drive the escape rate low enough on its own. **Second, the judge shares the generator’s blind spots.** An LLM judge asked whether an answer is grounded is itself a language model, prone to the same plausibility bias that produced the error; when retrieval has fetched a confident-sounding but wrong passage, both generator and judge may agree on a falsehood. **Third, confidence is usually self-reported.** Many systems surface the model's own stated certainty as a quality signal, but neural networks are poorly calibrated[6] and a fabricated value frequently carries high confidence. Trusting a model's self-assessment of whether it hallucinated is circular.

The conclusion is not that judges are useless but that no single mechanism is sufficient. What reduces the escape rate is *independent* layers whose failure modes do not coincide: grounding that limits what can be said, determinism that limits where the model runs, verification by signals that do not all share the generator's biases, abstention that converts uncertainty into a safe default, and monitoring that catches what slips through over time. Table 1 frames the contrast.

**Table 1. A single faithfulness judge versus a layered containment architecture.**

| **Property** | **RAG + single judge** | **HALO (layered)** |
|---|---|---|
| Escape rate at volume | Bounded by one detector's recall | Product of independent layers' misses |
| Correlated blind spots | Judge shares generator's biases | Evidence-based checks break the correlation |
| Confidence signal | Often model self-reported | Verified against source text |
| Behavior when unsure | Usually still answers | Abstains or escalates by default |
| Regressions over time | Undetected until incident | Drift-detected, alerted, corrected |

## 3 The HALO Architecture

HALO arranges six layers around the generative core, each addressing a failure mode the others cannot fully close. A request is grounded before it is answered, generated under constraints, verified by independent signals, and either delivered with its evidence or withheld; everything is traced, and the whole is watched over time. Figure 1 shows the arrangement.

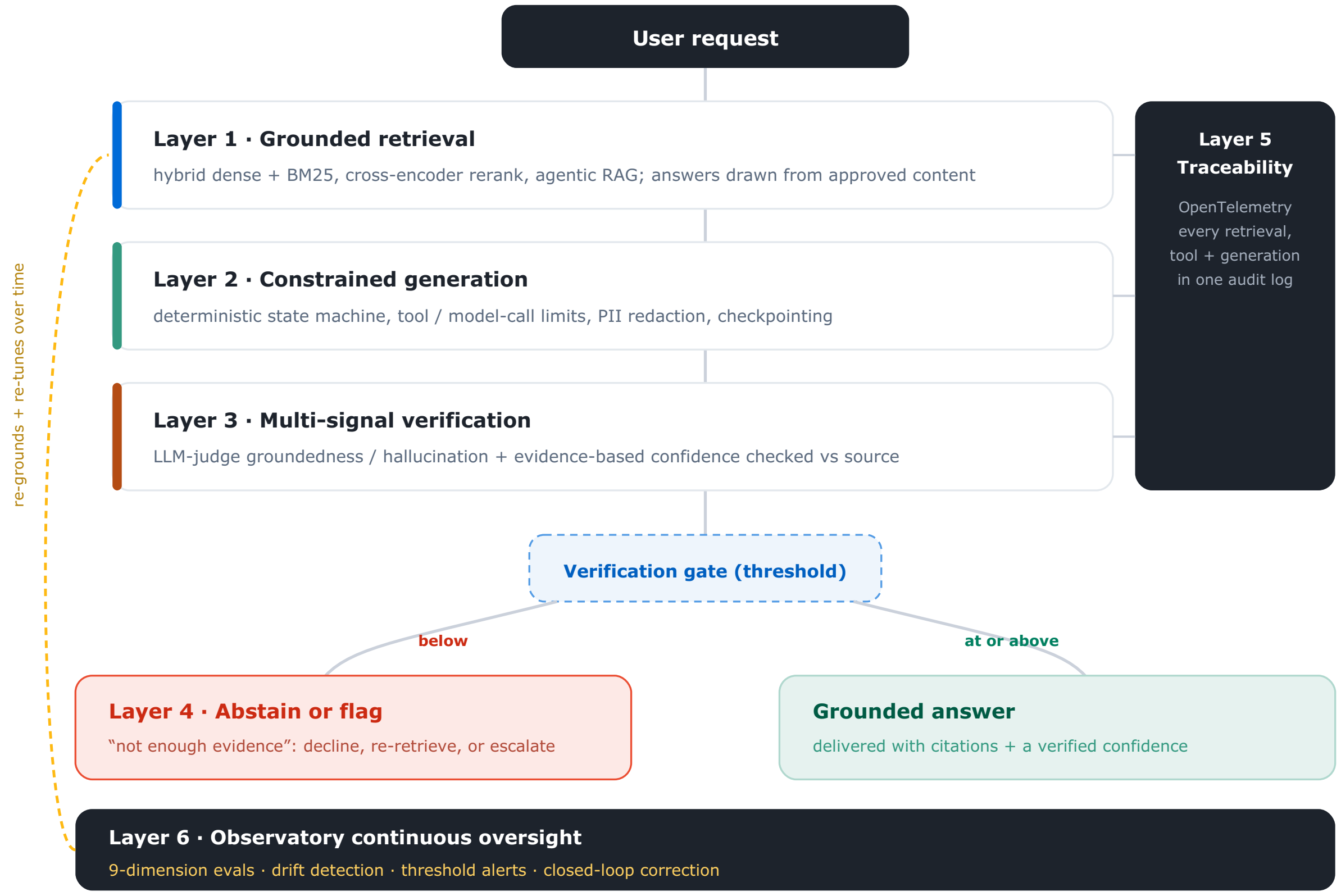

**Figure 1.** Each layer closes a gap the others leave open. Grounding limits what can be said; constrained execution limits where the model can run; multi-signal verification scores the output by independent checks; the gate routes anything under threshold to abstention rather than to the user; traceability records every step; and continuous oversight watches the whole over time, feeding regressions back into re-grounding and re-tuning. The escape rate is the product of the layers' individual misses, not any single detector's.

### 3.1 Layer 1: grounded generation

The first defense is to limit what the model is even asked to produce from memory. The platform answers from a knowledge base of approved, indexed content rather than parametric recall: documents are semantically chunked, embedded, and stored in a vector backend, then retrieved at query time. Retrieval is hybrid (dense vectors combined with sparse BM25) and two-stage, with a cross-encoder reranker promoting the genuinely relevant passages over the merely similar. For harder questions an agentic retrieval mode decomposes the query into sub-queries, retrieves for each, grades the candidates for relevance to the original question, and reformulates and retries when nothing relevant is found. Grounding does not by itself guarantee faithfulness, since a model can still misread or over-reach on good context, but it converts an open-ended generation problem into a bounded one: the answer should be entailed by retrieved, approved text, and that claim is check-able.

### 3.2 Layer 2: constrained execution

The second defense bounds where the model can go wrong. Agent workflows run as deterministic state machines with checkpointing, not as open-ended autonomous loops, so the set of reachable states is known and replayable. Middleware enforces hard limits on tool calls and model calls, redacts personally identifiable information before it reaches a provider, and applies node-level guards. The effect is to shrink the blast radius: a model that misbehaves on one node cannot silently rewrite the workflow, exhaust a budget, or leak data, and an operator can reconstruct exactly what happened. Determinism around a stochastic core is what makes the rest of the architecture auditable.

### 3.3 Layer 3: multi-signal verification

The third defense scores every output for support, using more than one independent signal so that their blind spots do not coincide. An LLM-as-judge evaluation rates groundedness, hallucination, and correctness against the retrieved context, in the lineage of atomic-claim faithfulness scoring[3],[4]. Crucially, this is complemented by *evidence-based* checks that do not run through a language model at all: for extracted data, the system verifies each value against the source document by exact, fuzzy, and label-proximity matching, and validates internal consistency (for example, that invoice line items sum to the stated total, or that an invoice date precedes its due date). Because these checks are deterministic and source-anchored, they catch the case an LLM judge is worst at, a confident fabrication that reads as plausible, and they break the correlation between generator and judge errors that undermines single-signal designs. Section 4 treats this layer in detail.

### 3.4 Layer 4: calibrated abstention

The fourth defense is the one most systems omit: a safe default when evidence is insufficient. When retrieval returns nothing relevant, or verification scores fall below threshold, HALO does not let the model improvise an answer; the gate routes the request to abstention. In practice this means the agent says it does not have enough information, escalates to a human, or re-retrieves, rather than emitting a fluent guess. This is the operational meaning of “if unsure, the system says so.” Abstention is unglamorous and occasionally frustrating to a user, but in a regulated setting a declined answer is almost always cheaper than a confident wrong one, and making refusal a first-class, monitored behavior is what lets the enterprise trust the answers that are delivered.

### 3.5 Layer 5: total traceability

The fifth defense assumes the others will occasionally fail and makes failure investigable. Every retrieval query, tool execution, and model call is captured through OpenTelemetry-native instrumentation into a single, complete audit log. When an answer is questioned, an operator can see exactly which passages were retrieved, which prompt was sent, what the model returned, and how each verification signal scored it. Traceability does not prevent a hallucination, but it converts “the AI said something wrong” from an unfalsifiable complaint into a specific, reproducible record, which is both an engineering necessity and, increasingly, a regulatory one[7],[8],[9].

### 3.6 Layer 6: continuous oversight

The sixth defense addresses time. A workflow that was faithful at launch can degrade as inputs shift, vocabularies change, or a provider updates a model. The oversight plane runs continuous evaluation across nine quality dimensions (including correctness, hallucination, and groundedness), performs statistical drift detection on multi-metric baselines, and raises alerts when thresholds are breached. Oversight is not merely passive: it closes the loop, feeding regressions into a correction cycle that regenerates candidate agent definitions and validates them before any reaches production. Section 5 describes that loop.

## 4 Evidence-Based Confidence

The verification layer’s most distinctive component, and the one that most directly separates HALO from a bolt-on judge, is its treatment of confidence. The question “how sure are we that this extracted value is real?” is answered not by asking the model, but by checking the value against the document it supposedly came from. Figure 2 shows the pipeline for a single extracted field.

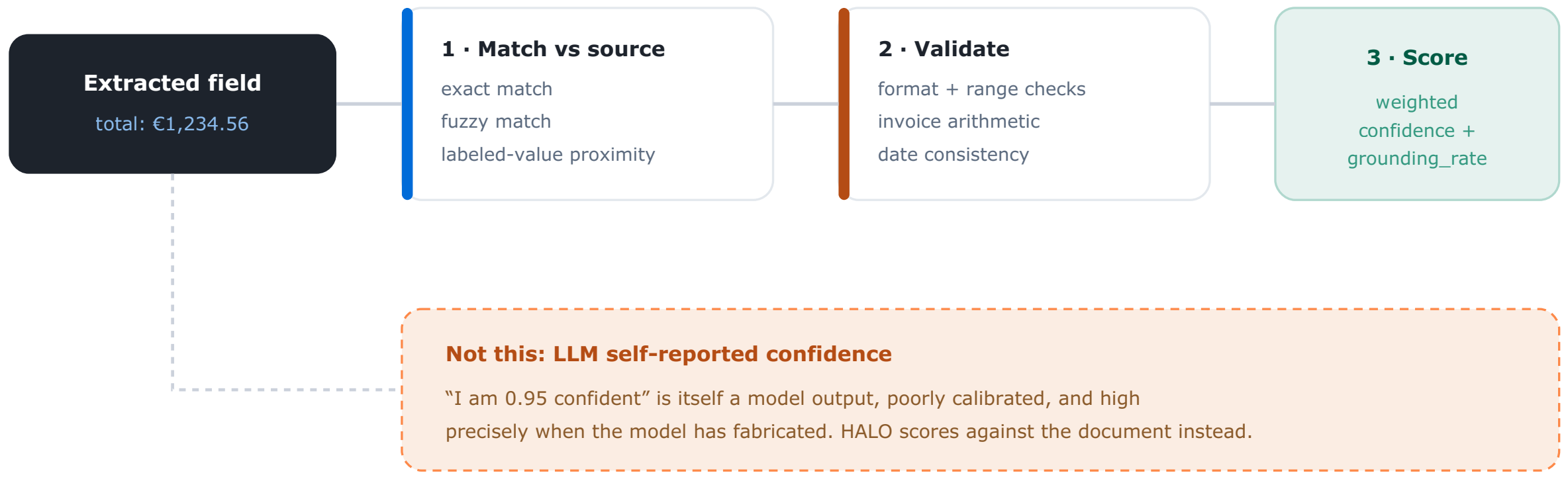

**Figure 2.** For each extracted field the toolkit locates the value in the source text (exact, fuzzy, or near an expected label), validates it structurally and against sibling fields, and produces a weighted confidence and an overall *grounding_rate*. None of these checks runs through the language model, so a confident fabrication, the case a model judge is weakest on, is exactly the case they catch.

### 4.1 Matching the value to the source

The first step locates the extracted value in the original document. An exact string match yields the highest confidence; where formatting differs, a fuzzy match scores by similarity ratio; and a labeled-value search looks for the value near an expected anchor such as "Total:" or "Amount Due:", which both confirms the value and confirms its meaning. A value that cannot be found in the source at all is the strongest possible hallucination signal, and it is invisible to a judge that only reads the answer.

### 4.2 Field and cross-field validation

The second step applies structure. Field-specific scorers combine weighted signals; an extracted amount, for instance, is scored on exact match, label proximity, format validity, and a plausibility range check, so that a value present in the document but in the wrong role scores lower than one found beside its proper label. Cross-field validation then checks internal consistency: line items should sum to the subtotal within tolerance, tax and total should reconcile, and an invoice date should precede its due date. These constraints catch a class of error no amount of reading fluency would, because they test the answer against arithmetic and logic rather than against plausibility.

### 4.3 From confidence to action

The third step aggregates the signals into a per-field confidence and an overall grounding rate, and flags the fields most at risk of being hallucinated. These scores are not cosmetic: they feed the verification gate of Section 3.4. A document whose key fields cannot be grounded does not yield a quietly-wrong extraction; it is routed to abstention and human review. The same evidence-based philosophy governs the upstream document parser, which grades OCR quality per page and escalates low-confidence pages (below an 85% threshold) to a stronger vision model rather than passing through a noisy read. At no point does the system rely on the model's opinion of its own accuracy.

## 5 Continuous Oversight and Self-Correction

Containment at request time is necessary but not sufficient, because quality moves. The oversight layer treats faithfulness as something to be monitored and maintained, not certified once. Every run is scored across nine evaluation dimensions, of which correctness, hallucination, and groundedness are the load-bearing ones for trust; the rest (tool use, RAG coverage, refusal, toxicity, conciseness, helpfulness) round out a quality profile.

These scores form multi-metric baselines, and statistical drift detection raises a signal when behavior shifts away from them, before a human notices an incident.

### 5.1 From signal to candidate fix

When a metric breaches a threshold, for example a hallucination rate climbing above its service-level objective, oversight closes the loop rather than merely paging an engineer. A correction cycle reads the offending traces and evaluations to localize the cause to a node, prompt, or tool, then generates a set of candidate fixes: a structured-output instruction, a post-extraction validation node, a temperature or model change, an added negative example. Each candidate is a hypothesis about how to restore faithfulness, and HALO does not deploy a hypothesis on intuition. It runs an experiment, in two stages, offline then online.

### 5.2 Offline evaluation: experiments on golden datasets

The first stage is a controlled offline experiment. Every candidate is executed against a curated golden dataset of representative cases with known expected outputs, and scored on the same nine dimensions used in production. Because the dataset, the baseline, and the scoring are fixed, the comparison is reproducible and fair: candidates differ only in the change under test. A candidate is allowed to proceed only if its improvement over the baseline is *statistically significant* (for example, passing a paired test at the 0.05 level on the trust dimensions) and its cost impact is acceptable. Offline evaluation is cheap, fast, and safe because no real user is exposed, and it is where most candidates are eliminated; it answers the question "is this change better on cases we already understand?" before any traffic is risked. It cannot, however, answer whether the change survives the messiness of live inputs, which is why it gates rather than decides.

### 5.3 Online evaluation: canary experiments in production

The second stage is an online experiment. The offline winner is not trusted on the benchmark alone; it is deployed behind a canary that takes a small share of live traffic (for instance, ten percent) while the incumbent serves the rest. The two arms are evaluated online with the same continuous scoring and drift detection applied to everything else, so the comparison is now against real, unseen inputs rather than a curated set. If the canary's production metrics hold or improve over a monitoring window, traffic shifts to it fully; if they degrade, it is rolled back automatically, and the incumbent is never disturbed. Offline evaluation decides what is *allowed* to ship; online evaluation confirms it *should*. The pairing matters because the two stages fail differently: a fix can look strong on a golden dataset and regress on the long tail of real documents, and only a live experiment will catch that before it becomes the new baseline.

### 5.4 What the loop does and does not promise

It is worth being precise. This loop does not make the model incapable of hallucinating. It ensures that when the rate of hallucination rises, the rise is measured, attributed, corrected by a change that is validated offline and then online, and prevented from silently reaching users in the interim. Trust, in HALO, is not a one-time property of a deployment but a continuously defended one, and the same evidentiary discipline the system applies to its answers it applies to its own improvement.

## 6 Illustrative Application

Consider a claims-extraction agent at an illustrative regulated insurer, Meridian Insurance, that reads submitted claim documents and extracts the structured fields a downstream workflow needs: claimant, policy number, incident date, and claimed amount. The figures here are illustrative and chosen to expose the mechanism, not to report a customer result.

In its first production weeks the agent looks acceptable on a casual review, but the oversight plane tells a different story. Across the nine-dimension evaluation it scores a hallucination rate of **0.28** and correctness of **0.72**, both outside the service-level objectives of 0.20 and 0.90, and drift detection grades the shift as high. The traces localize the problem to a single node, the claim-amount extraction, and the evidence-based confidence toolkit is precise about why: on a meaningful fraction of documents the extracted amount cannot be matched to the source text, and the cross-field arithmetic does not reconcile. These are not answers a self-reported confidence would have questioned; the model returned several of the wrong amounts with high stated certainty.

Two things happen, and the distinction is the whole point of HALO. First, *at request time*, the ungrounded extractions never reach the downstream workflow: the verification gate routes each low-grounding document to abstention and human review rather than emitting a fabricated amount. The hallucination is contained before it becomes a mis-paid claim. Second, *over time*, the oversight loop corrects the cause. It generates candidate fixes for the offending node and evaluates them on a fifty-case golden dataset; the winning candidate, which adds a structured-output instruction and a post-extraction validation step, lifts correctness to **0.88** and cuts hallucination to **0.08**, a statistically significant improvement at an acceptable cost increase, and is promoted behind a canary with automatic rollback.

The agent still runs on a model that can hallucinate. What changed is that the system around it grounded the answer, verified it against the source, refused to pass the unverifiable ones, traced every step for the eventual audit, and then measured the regression and repaired it. "Zero hallucination" was never a claim about the model; it was the behavior of the harness, and the harness held.

## 7 Empirical Evaluation

The illustrative case study above is deliberately hypothetical. To give the grounding mechanism of Section 4 an actual, independently checkable result, we built a standalone reference implementation of its central claim, exact and fuzzy matching of extracted values against a source, plus cross-field arithmetic, aggregated into a weighted grounding score, and evaluated it on a public hallucination benchmark. This is not FlowX's production toolkit, which additionally combines an LLM-judge signal and is not available outside the company; it is a minimal, reproducible reimplementation built to test the mechanism this paper describes, and the code and test suite are released as supplementary material.

### 7.1 Dataset and setup

We use RAGTruth[14], a corpus of real LLM responses (GPT-4, GPT-3.5, Mistral, Llama-2) to question-answering, summarization, and data-to-text generation tasks, with human annotations marking spans that evidently conflict with or are baseless relative to the provided source. We treat a response as hallucinated if it carries at least one such annotation. The corpus splits into 15,090 training and 2,700 test responses, evenly divided across the three task types. The grounding-score abstention threshold is calibrated once on the training split (maximizing hallucination-detection F1) and then applied, unchanged, to the held-out test split; all numbers below are out-of-sample. We compare the full mechanism (exact + fuzzy + cross-field arithmetic matching) against a verbatim-only baseline that uses the identical claim-extraction step but accepts only exact substring matches, isolating what the fuzzy and arithmetic signals contribute.

### 7.2 Detection results

Table 1 reports hallucination-detection quality on the full test split. The multi-signal grounding score modestly outperforms verbatim-only matching (F1 0.556 vs. 0.551, AUROC 0.640 vs. 0.627), but the aggregate masks a sharper, more informative pattern in Table 2: the mechanism is a strong signal exactly where Section 4 designed it to be, structured field verification against a source record (Data2txt, F1 = 0.776, AUROC = 0.667), and a materially weaker one on free text without an explicit field structure (QA, Summary; AUROC 0.569–0.575). This is the honest shape of the claim: source-anchored matching is not a general-purpose hallucination detector, it is a strong detector for the class of extraction workloads HALO targets, and Section 8 says so explicitly.

**Table 1. Hallucination detection on RAGTruth test (n=2,700), threshold calibrated on train.**

| System | Precision | Recall | F1 | AUROC | Avg. Precision |
|---|---|---|---|---|---|
| HALO grounding (exact+fuzzy+arithmetic) | 0.415 | 0.842 | **0.556** | **0.640** | 0.416 |
| Verbatim-match baseline | 0.405 | 0.865 | 0.551 | 0.627 | 0.404 |

**Table 2. HALO grounding score by task type (n=900 each).**

| Task type | Precision | Recall | F1 | AUROC |
|---|---|---|---|---|
| Data2txt (structured record → text) | 0.654 | 0.952 | **0.776** | **0.667** |
| Summary (free text) | 0.266 | 0.475 | 0.341 | 0.575 |
| QA (free text) | 0.207 | 0.912 | 0.338 | 0.569 |

## 7.3 Calibrated abstention

Table 3 sorts test responses by grounding score and simulates abstaining on the least-grounded fraction, reporting accuracy (fraction non-hallucinated) among the responses retained. Abstention trades coverage for reliability in the direction the architecture intends: declining the least-grounded half of responses raises retained accuracy from 65.1% to 76.0% under the full mechanism, versus 75.0% under verbatim-only matching, a small but consistent edge for the multi-signal score as the operating threshold tightens.

**Table 3. Risk-coverage: accuracy among retained (non-abstained) responses.**

| Abstain fraction | Coverage | Retained accuracy (HALO) | Retained accuracy (baseline) |
|---|---|---|---|
| 0% | 100% | 65.1% | 65.1% |
| 20% | 80% | 67.4% | 67.2% |
| 50% | 50% | **76.0%** | 75.0% |

## 7.4 Adding the LLM-judge signal, and comparing models

Section 7.2–7.3 measured evidence-based matching alone. Section 3.3 also specifies an LLM-judge signal, so we added one, run independently on the same RAGTruth test responses by two different models (Claude Haiku 4.5 and GPT-5.4-mini), and combined each judge's risk score with the rule-based grounding score by simple averaging. Table 4 reports ranking quality (AUROC, average precision; threshold-free, so no train/test split is needed for this sweep) for all three signals.

**Table 4. Rule-based, judge-alone, and combined detection quality on RAGTruth test (n=2,700).**

| Signal | AUROC | Avg. Precision |
|---|---|---|
| Rule-based grounding alone (Table 1) | 0.640 | 0.416 |
| LLM judge alone (Claude Haiku 4.5) | **0.846** | 0.681 |
| LLM judge alone (GPT-5.4-mini) | 0.843 | 0.694 |
| Combined: rule + Claude judge (averaged) | 0.841 | 0.692 |
| Combined: rule + GPT judge (averaged) | 0.837 | **0.705** |

Two findings are worth stating plainly, including the one that complicates the architecture's simplest pitch. First, the judge signal is substantially stronger than the rule-based check on its own (AUROC 0.84–0.85 vs. 0.64), and the two judge models agree closely (0.846 vs. 0.843): the finding is not sensitive to which frontier model implements the judge, which is the multi-model result Section 3.3's design should produce if the judge signal is doing real work rather than encoding one model's idiosyncrasies. Second, naively averaging the rule-based and judge scores does *not* beat the judge alone here (0.841 and 0.837 vs. 0.846 and 0.843): a weaker signal, simply averaged in, pulls the ranking down slightly rather than up. This does not refute the multi-signal design, independent signals with different blind spots are still the right architecture for containment, but it does mean the combination rule matters: a naive average is the wrong one, and a real deployment should combine signals by a method that lets the stronger signal dominate (for example, a learned weight, or an OR-gate on independent abstention thresholds) rather than an unweighted mean. We report this rather than the more flattering alternative because the paper's claim should rest on what composition actually buys, not on which composition was tried first.

### 7.5 What this evaluation does and does not show

This experiment isolates the evidence-based matching signal of Section 4 and the judge signal of Section 3.3 in a standalone, reproducible harness; it does not exercise the OCR-escalation path or the online-canary loop of Section 5, neither of which is testable without the production system. A full validation of all six layers together, with a learned (not averaged) signal combination, remains future work. The reference implementation, the RAGTruth evaluation script, the judge sweep, and a pytest suite that reproduces Tables 1–4 are released as supplementary material.

## 8 Discussion

Framing zero hallucination as a system property rather than a model property changes what an enterprise should ask of an AI vendor. The useful questions are no longer "does your model hallucinate?" (every model can) but "what grounds the answer, what independent signals verify it, what does the system do when it is unsure, can you show me the trace, and how would you know if it got worse?" These are answerable, testable, auditable questions, and they map directly onto HALO's layers. They also generalize: the architecture is not specific to insurance or to document extraction. Any workflow where an agent produces claims that can, in principle, be checked against a source admits the same treatment, with only the grounding corpus, the validation rules, and the evaluation dataset changing.

The approach aligns naturally with where assurance and regulation are heading. Emerging AI-management standards and high-risk-AI regulation increasingly demand exactly what a containment architecture produces as a byproduct: documented grounding, recorded verification, and a traceable account of how each output was produced and checked[7],[8],[9]. An organization that contains hallucination this way is, almost incidentally, generating the evidence its auditors and regulators will ask for.

## 9 Limitations

- **The empirical evaluation covers two layers, not six.** Section 7 measures the evidence-based grounding signal and the LLM-judge signal, independently and combined by simple averaging, on a public benchmark, without the OCR-escalation path or the online-canary loop; it demonstrates that the evidence-based signal carries real signal on structured extraction and is markedly weaker on free text, that the judge signal is stronger and consistent across two model providers, and that naive averaging is the wrong way to combine them, not that the full six-layer architecture achieves any particular hallucination rate in production.
- **Containment is not elimination.** HALO reduces the rate at which hallucinations reach users toward zero by composing independent layers; it does not make the underlying model incapable of error, and a residual escape rate remains, bounded by the product of the layers' individual misses.
- **Grounding requires a checkable source.** Evidence-based verification is strongest where the answer can be matched against a document or computed constraint. For genuinely generative or judgment tasks with no source of truth, the architecture falls back on the LLM judge and abstention, which are weaker guarantees.
- **Abstention has a cost.** Declining or escalating when evidence is thin trades coverage for safety. Tuning the verification threshold is a real decision with a real false-abstention rate, and the right operating point is domain-specific.
- **The judge layer shares model biases.** The LLM-as-judge signal remains correlated with the generator; HALO mitigates this with the evidence-based and consistency signals, but on tasks where those do not apply, that correlation is not fully broken.
- **Oversight depends on representative evaluation data.** Drift detection and the correction loop are only as good as the baselines and golden datasets behind them; an unrepresentative dataset can hide a regression or promote a fix that does not generalize.

## 10 Conclusion

The enterprise demand for "AI that does not hallucinate" is reasonable in spirit and impossible to meet at the level of the model. The productive response is to stop trying to perfect the component and start engineering the system: ground generation in approved content, constrain where the model runs, verify every output by signals that do not share the generator's blind spots, abstain rather than guess when the evidence is thin, trace everything, and watch the whole over time so that regressions are caught and corrected. HALO is that system. It does not promise a model that never errs. It promises something an enterprise can actually rely on and an auditor can actually inspect: a harness in which a hallucination cannot reach the user unseen, unflagged, or uncorrected. That, and not a mythical infallible model, is what "zero hallucination" should mean.

## Appendix A  The Nine Evaluation Dimensions

Every run is scored across nine dimensions. The first three are the load-bearing trust signals; the remainder round out a quality profile and feed the composite used by the correction loop.

| Dimension | Question it answers | Role in HALO |
|---|---|---|
| **Groundedness** | Is every claim supported by the retrieved context? | Primary gate signal |
| **Hallucination** | Does the output assert anything unsupported? | Primary gate + SLO |
| **Correctness** | Is the answer right against expected output? | Primary SLO |
| RAG coverage | Did retrieval surface the needed evidence? | Diagnoses grounding failures |
| Tool use | Were tools selected and called appropriately? | Diagnoses execution faults |
| Refusal | Did the agent abstain when it should have? | Validates Layer 4 |
| Toxicity | Is the output free of harmful content? | Safety guard |
| Conciseness | Is the answer free of padding? | Quality profile |
| Helpfulness | Does it actually address the request? | Quality profile |

## Appendix B  Worked Example: Scoring One Extracted Field

This appendix scores a single extracted field against its source, the arithmetic of Section 4 made concrete. The figures are illustrative.

### *B.1  The extraction*

From a claim document, the agent extracts *total = €1,234.56*. The evidence-based toolkit scores it on four weighted signals rather than asking the model how sure it is.

1. **Exact match** (weight 0.30). The string “1,234.56” is found verbatim in the source text: signal = 1.0.

2. **Labeled value** (weight 0.40). The value appears immediately after the anchor “Total Due:”, confirming both the figure and its role: signal = 1.0.

3. **Format validation** (weight 0.15). The value parses as a well-formed currency amount: signal = 1.0.

4. **Range check** (weight 0.15). The amount falls within the plausible range for this document class: signal = 1.0.

The weighted confidence is *0.30(1.0) + 0.40(1.0) + 0.15(1.0) + 0.15(1.0) =* ***1.00***. The field is grounded and passes the gate.

### *B.2  A hallucinated counterpart*

Now suppose the model instead returns *total = €1,243.56* (two digits transposed). The exact-match and labeled-value searches fail to find this string near “Total Due:” (signals near 0.0); only the format and range checks pass. The weighted confidence collapses to roughly *0.30(0) + 0.40(0) + 0.15(1.0) + 0.15(1.0) =* ***0.30***, and cross-field validation compounds the signal when the line items no longer sum to the stated total. The field is flagged, the document is routed to abstention, and the transposition never reaches the downstream workflow. A self-reported confidence would likely have returned a high value for both extractions; the source-anchored score separates them cleanly.

## Appendix C  Containment Matrix: Failure Mode to Layer

Defense-in-depth means each hallucination failure mode is caught by at least one independent layer, and usually more than one. The matrix below maps common failure modes to the primary layer that contains them.

| Failure mode | Primary containment | How it is caught |
|---|---|---|
| Fabricated fact from model memory | L1 Grounding | Answer must come from retrieved approved content |
| Over-reach on good context | L3 Verification | Groundedness / atomic-claim scoring vs context |
| Confident fabricated value | L3 Evidence-based | Value not matchable to source text |
| OCR misread of a figure | L3 + parser | Per-page OCR grade; VLM fallback below 85% |
| Internally inconsistent extraction | L3 Cross-field | Arithmetic / date-consistency validation |
| Answer with no sufficient evidence | L4 Abstention | Gate routes below-threshold to refusal / escalation |
| Runaway tool / model behavior | L2 Constraints | Deterministic state machine + call limits |
| Quality regression over time | L6 Oversight | Drift detection + closed-loop correction |
| Disputed answer after the fact | L5 Traceability | Full OpenTelemetry record of the run |